\documentclass[final,1p,times]{elsarticle}

\usepackage{amssymb}
\usepackage{amsmath,amsfonts}
\usepackage{algorithmic}
\usepackage{algorithm}
\usepackage{color,array}
\usepackage[caption=false,font=normalsize,labelfont=sf,textfont=sf]{subfig}
\usepackage{textcomp}
\usepackage{stfloats}
\usepackage{url}
\usepackage{verbatim}
\usepackage{graphics}
\usepackage{multirow}
\usepackage{bm}
\usepackage{threeparttable}
\usepackage{booktabs}
\usepackage{subfloat}
\usepackage{makecell}

\newtheorem{myDef}{Definition}

\journal{}

\biboptions{authoryear}

\begin{document}

\begin{frontmatter}

\title{Heterogeneous Graph Contrastive Learning with Spectral Augmentation}


\author[mymainaddress]{Jing Zhang \corref{mycorrespondingauthor}}
\ead{jingz@seu.edu.cn}

\author[mymainaddress]{Xiaoqian Jiang}
\ead{220224926@seu.edu.cn}

\author[secondaddress]{Yingjie Xie}
\ead{yjx0921@njust.edu.cn}

\author[secondaddress]{Cangqi Zhou}
\ead{cqzhou@njust.edu.cn}

\cortext[mycorrespondingauthor]{Corresponding author}
\address[mymainaddress]{School of Cyber Science and Engineering, Southeast University, No. 2 SEU Road, Nanjing 211189, China}
\address[secondaddress]{School of Computer Science and Engineering, Nanjing University of Science and Technology, 200 Xiaolingwei Street, Nanjing 210094, China}

\begin{abstract}
	Heterogeneous graph can well describe the complex entity relationships in the real world. For example, online shopping networks contain multiple physical types of consumers and products, as well as multiple relationship types such as purchasing and favoriting. More and more scholars pay attention to this research because heterogeneous graph representation learning shows strong application potential in real world scenarios. However, the existing heterogeneous graph models use data augmentation techniques to enhance the use of graph structure information, which only captures the graph structure information from the spatial topology, ignoring the information displayed in the spectrum dimension of the graph structure. To address the issue that heterogeneous graph representation learning methods fail to model spectral information, this paper introduces a spectral-enhanced graph contrastive learning model, SHCL, and proposes a spectral augmentation algorithm for the first time in heterogeneous graph neural networks. This model learns an adaptive topology augmentation scheme through the heterogeneous graph itself, disrupting the structural information of the heterogeneous graph in the spectrum dimension, and ultimately improving the learning effect of the model. Experimental results on multiple real-world datasets demonstrate the model’s substantial advantages.
\end{abstract}


\end{frontmatter}


\section{Introduction} \label{Sec:Intro}
As a complex network structure that can represent various types of nodes and edges, heterogeneous graph can intuitively model most complex scenes in reality, such as user-blog network in blog platform, user-commodity network in e-commerce platform and protein network in biochemistry. Benefiting from the advantages of intuitively modeling realistic scenes, heterogeneous graphs are widely used in various fields.

In recent years, many scholars have proposed various methods to mine and analyze the rich semantic information in heterogeneous graphs, namely heterogeneous graph representation learning methods \citep{shi2022heterogeneous}. For example, \cite{zhang2019heterogeneous} proposed HetGNN, which is capable of learning fusion embeddings from multimodal node properties (such as images, text, or speech) using attention mechanisms. \cite{ren2019heterogeneous} extended the idea of infomax to heterogeneous graphs and used self-supervised signals to learn node information in heterogeneous graphs. And \cite{wang2021self} proposed HeCo, which can capture both meta-path \citep{dong2017metapath2vec} information and network schema \citep{zhao2020network} information by using the collaborative comparison mechanism. Under semi-supervised conditions, \cite{wang2019heterogeneous} proposed a heterogeneous graph attention network (HAN), which uses a hierarchical attention mechanism to capture the importance of nodes and semantics. However, these methods still have two challenges to solve in the process of mining the rich information in heterogeneous graphs.
\begin{itemize}
	\item \textbf{The lack of exploring the essential information in heterogeneous graphs.} Most studies uniformly model various node attributes and complex heterogeneous structures in heterogeneous graphs without distinguishing which information is the most critical information in heterogeneous graph representation learning \citep{wang2022survey}. These information, to some extent, represent the most essential features of heterogeneous graphs (or nodes) that are different from other heterogeneous graphs (or nodes). Compared with homogeneous graphs, heterogeneous graphs contain more types of nodes and more complex special structures, so it is more critical to explore the essential information of heterogeneous graphs. Without a clear understanding of this essential information, there will be no direction when studying all aspects of heterogeneous graphs, and some unimportant or even irrelevant information may be introduced, which will ultimately affect the modeling effect of the model on heterogeneous graphs.
	\item \textbf{Lack of multidimensional cognition of heterogeneous graph structure.} Most of the existing methods only model the semantic structure information of heterogeneous graphs from a single spatial topological dimension \citep{yang2023simple}, ignoring the important information displayed by the graph structure in the spectrum dimension. For example, the eigenvalues and eigenvectors of the Laplacian matrix of a graph can reflect the degree of similarity of node properties. \citep{dong2016learning} If the heterogeneous graph information is mined only from the spatial topological dimension, these node association features will be ignored.
\end{itemize}

In order to solve these challenges, we propose a novel heterogeneous graph contrastive learning model SHCL based on spectral augmentation algorithm. The main goal of SHCL is to learn a topology augmentation scheme specific to the heterogeneous graph according to its own characteristics, which can disturb the spectral information of the heterogeneous graph to the maximum extent. The heterogeneous graph encoder, in turn, learns the underlying spectral invariance from these views with vastly different spectrums. Specifically, SHCL begins with a spectral augmentation objective function that continuously updates the parameterized topology augmentation scheme. Secondly, these topological augmentation schemes are performed on heterogeneous graphs, and the enhanced views with great difference between the two spectra are obtained. A dual aggregation (network schema and meta-path) encoder is then used to learn the spectral invariance of the heterogeneous graph.

Overall, our key contributions are summarized as follows:
\begin{itemize}
	\item To solve the problem that the heterogeneous graph representation learning method can not model the spectrum information, we propose a heterogeneous graph contrastive learning model SHCL based on spectral augmentation algorithm.
	\item The SHCL model learns the adaptive topology augmentation scheme through the information of the heterogeneous graph itself, disturbs the structural information of the heterogeneous graph in the spectrum dimension, and finally improves the performance of the model in the downstream task. 
	\item We conducted experiments on multiple real-world datasets to confirm the significant advantages of SHCL.
\end{itemize}
\section{Related Work}\label{sec:rwk}
\subsection{ Graph Contrastive Learning}
Current graph contrastive learning models typically train encoders to differentiate between positive and negative samples, thereby learning representations of nodes or graphs. Specifically, \cite{hassani2020contrastive} proposed to learn node-level and graph-level representations by manipulating node diffusion and comparing node representations with augmented graph representations. \cite{qiu2020gcc} proposed a pre-training framework based on contrastive learning. It proposes to obtain subgraphs through random walks, conduct random sampling to build multiple views, and then learn model weights using multiple feature schemes. In contrast, \cite{you2020graph} discuss graph contrastive learning from the perspective of the invariance of the execution disturbance. They propose a data-augmented Contrastive Learning Framework (GraphCL) for learning unsupervised representations of graph data. \cite{zhu2021graph} proposed an adaptive augmentation method for the random augmentation method of GraphCL, which can select the attributes and edges of disturbed nodes according to the characteristics of nodes or edges. \cite{kalantidis2020hard} proposed a feature-level hard-to-negative hybrid strategy that forces the model to learn more robust features by synthesizing novel examples, and computes them instantaneously with minimal computational overhead.
\subsection{Heterogeneous graphs represent learning}
Heterogeneous graph representation learning aims to learn functions that map input Spaces to low-dimensional Spaces, taking into account different aspects of information, including graph structure, properties, and application-specific labels. According to the information used by existing methods in heterogeneous graph representation learning, it is divided into three types: structure-preserving heterogeneous graph representation learning methods, attribute-assisted heterogeneous graph representation learning methods and application-oriented heterogeneous graph representation learning methods.

\textit{Structure-preserving heterogeneous graph representation learning methods.} Methods that fall into this category primarily focus on capturing and preserving heterogeneous structures (such as meta-path and meta-gram) and underlying semantics. For example, the PME
 \citep{chen2018pme} treats each edge type as a relation and uses relationship-specific matrices to transform nodes into different metric Spaces. metapath2vec \citep{dong2017metapath2vec} uses meta-path-guided random walks to generate sequences of heterogeneous nodes with rich semantics. metagraph2vec \citep{zhang2018metagraph2vec} uses meta-graph-guided random walks to generate sequences of heterogeneous nodes. Unlike metagraph2vec, which only uses meta-graphs in the pre-processing step (i.e. using meta-graph to guide random walks), mg2vec \citep{zhang2020mg2vec} co-learns meta-graph and node embedding in order to add meta-graph to the representation learning process.
 
 \textit{Attribute-assisted heterogeneous graph representation learning methods.} Methods that fall into this category mainly focus on incorporating more information outside the structure (such as node attributes and edge attributes) into heterogeneous graph representation learning, thereby making more efficient use of neighborhood information to learn node embeddings. For example, the Heterogeneous graph Attention Network (HAN) \citep{wang2019heterogeneous} uses a hierarchical attention mechanism to capture the importance of nodes and semantics. \cite{fu2020magnn} further considered the intermediate nodes of the meta-path and proposed the MAGNN model. On this basis, GTN \citep{yun2019graph} was proposed to automatically identify useful connections between unconnected nodes. In addition, HGT \citep{hu2020heterogeneous} is designed for network scale heterogeneous graphs through heterogeneous small-batch graph sampling algorithm.
 
 \textit{Application-oriented heterogeneous graph representation learning methods.} Methods that fall into this category primarily focus on integration with a few specific applications. For example, HERec \citep{shi2018heterogeneous} learns the representation vectors of users and projects under different meta-paths and merges them to make recommendations. Camel \citep{zhang2018camel} considers content information (such as paper text) and contextual information (such as co-occurrence of paper and author). \cite{liu2018subgraph} proposed a subgraph enhanced heterogeneous graph embedding method, which uses stacked autoencoders to learn subgraph embedding, thereby enhancing the effect of semantic proximity search.
 
 However, the above methods either do not take into account the extension of graph augmentation to heterogeneous graphs, or ignore the information that the graph structure presents in the spectral dimension.
 
\section{Preliminary}
In this section, we formally present several concepts that will be utilized throughout the subsequent sections of this paper.
\begin{myDef}
	\textit{\textbf{Heterogeneous Graphs.} A graph consisting of entities of different types (i.e., nodes) and/or relationships of different types (i.e., edges) can be defined as follows:}
	\begin{equation}
		\mathcal{G}=(\mathcal{V},\mathcal{E},\mathcal{X},\mathcal{O},\mathcal{R}) 
	\end{equation}	
\textit{where $\mathcal{V}$ is the set of nodes, $\mathcal{E}$ is the set of edges, $\mathcal{O}$ is the set of types of nodes, $\mathcal{R}$ is the set of types of edges, and $\mathcal{X}=\{X_1,...,X_{|\mathcal{O}|}\}$ is the set of feature matrices. Existing node type mapping function $\phi:\mathcal{V}\to\mathcal{O}$, edge type mapping function $\psi:\mathcal{E}\to\mathcal{R}$. And the heterogeneous graph satisfies the constraint  $|\mathcal{O}+\mathcal{R}|>2$, and accordingly, if  $|\mathcal{O}|=1$ and $|\mathcal{R}|=1$ , then the graph is a homogeneous graph.}
\begin{figure}
	\centering
		\includegraphics[width=5.4in]{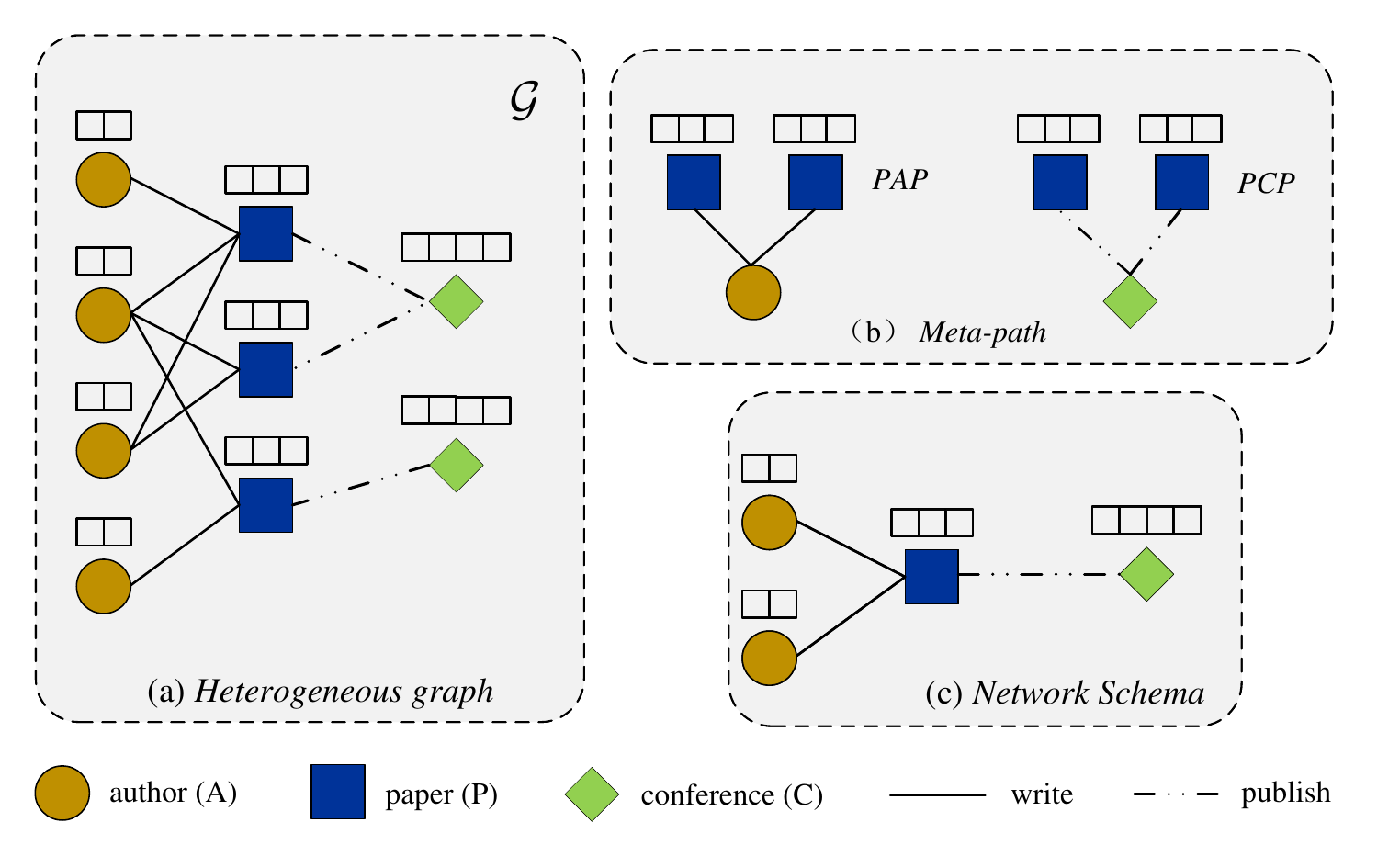}
	\caption{Examples and related concepts of heterogeneous graphs.}
	\label{fig1}
\end{figure}
\textit{Heterogeneous graphs not only provide the graph structure of data association, but also provide a higher level of data semantics. An example of a heterogeneous graph is shown in Figure~\ref{fig1} (a), which consists of three types of nodes (author, paper, and conference) and two types of edges (author-writer-paper, conference-publish-paper).}
\end{myDef}

\begin{myDef}
	\textit{\textbf{Meta-path.} Based on the constructed heterogeneous graph, \cite{shi2018heterogeneous} further proposed Meta-path in order to mine the semantics of higher-order relationships between entities.Meta-path $\mathcal{P}$ is a path defined on a heterogeneous graph and can be formally defined as:}
	\begin{equation}
		\mathcal{P}=O_1\stackrel{R_1}{\longrightarrow}O_2\stackrel{R_2}{\longrightarrow}...\stackrel{R_l}{\longrightarrow}O_{l+1}
	\end{equation}	
	\textit{Where $O_i\in\mathcal{O}$, $R_i\in\mathcal{R}$, $l$ is the length of the meta-path $\mathcal{P}$. $R=R_1\cdot R_2\cdot...\cdot R_l$ is a composite relationship between node types $O_1$ and $O_{L+1}$. As shown in Figure~\ref{fig1} (b), different meta-paths describe semantic relationships in different views. For example, the meta-path "APA" indicates a relationship where authors publish papers together, and "APCPA" indicates a relationship where authors attend conferences together. They can all be used to express the correlation between authors.}
\end{myDef}

\begin{myDef}
	\textit{\textbf{Network Schema.} While meta-path can be used to describe semantic correlations between entities, it cannot capture the heterogeneity of local structures. Network Schema is a heterogeneous structure defined on node types and edge types, and the mathematical definition can be formalized as follows:}
	\begin{equation}
		\mathcal{S}_{\mathcal{G}}=(\mathcal{O},\mathcal{R})
	\end{equation}	
	\textit{As shown in Figure~\ref{fig1} (c), the nodes of the Network Schema are the node types in the heterogeneous graph $\mathcal{G}$, and the relationships between the nodes (i.e., edges) are the edge types in the heterogeneous graph $\mathcal{G}$. The network pattern is used to describe the direct connections between different nodes and represents the local structure of the heterogeneous graph.}
\end{myDef}

\begin{myDef}
	\textit{\textbf{Graph Representation Learning.} The goal of graph representation learning is to generate graph/node representation vectors that can accurately capture the structure and node features of graphs. The performance of the graph representation learning model is measured by the performance of these vectors in downstream tasks such as node classification, graph classification, link prediction, and anomaly detection. Given a graph $G$, the node representation learning goal is to train an encoder:}
	\begin{equation}
		f_\theta:G\to \mathbb{R}^{n\times d'}	\end{equation}	
	\textit{Causes the encoder $f_\theta(G)$ to generate a low-dimensional vector for each node in $G$ that can be used for downstream tasks.And the graph representation vector can be further obtained by the Readout Function $g_\varphi:\mathbb{R}^{n\times d'}\to \mathbb{R}^{d'}$ to assemble the node representation vector set sum, so that the $g_\varphi (f_\theta(G))$ output a low-dimensional vector of the graph $G$, which can be used in downstream tasks at the graph level. $\Theta$ is used to represent all model parameters.}
\end{myDef}

\begin{myDef}
	\textit{\textbf{Graph Spectrum.} Graph representation learning can be viewed as Graph Signal Processing (GSP) \citep{lin2022spectral}. The Graph Shift Operator in GSP usually takes the normalized Laplacian matrix $L_{norm}$ and needs to decompose its features into:}
	\begin{equation}
		L_{norm}=UAU^\top
	\end{equation}	
	\textit{The diagonal matrix $\Lambda=eig(L_{norm})=diag(\lambda_1,...,\lambda_n)$ consists of real eigenvalues, which are called the Graph Spectrum. Accordingly, $U=[U_1,...U_n]\in\mathbb{R}^{n\times n}$ consists of orthogonal eigenvectors, which are called Spectral Bases.}
\end{myDef}

\section{The Proposed Method}\label{sec:mtd}

\subsection{Overview of the Model}
\begin{figure}
	\centering
	\includegraphics[width=5.34in]{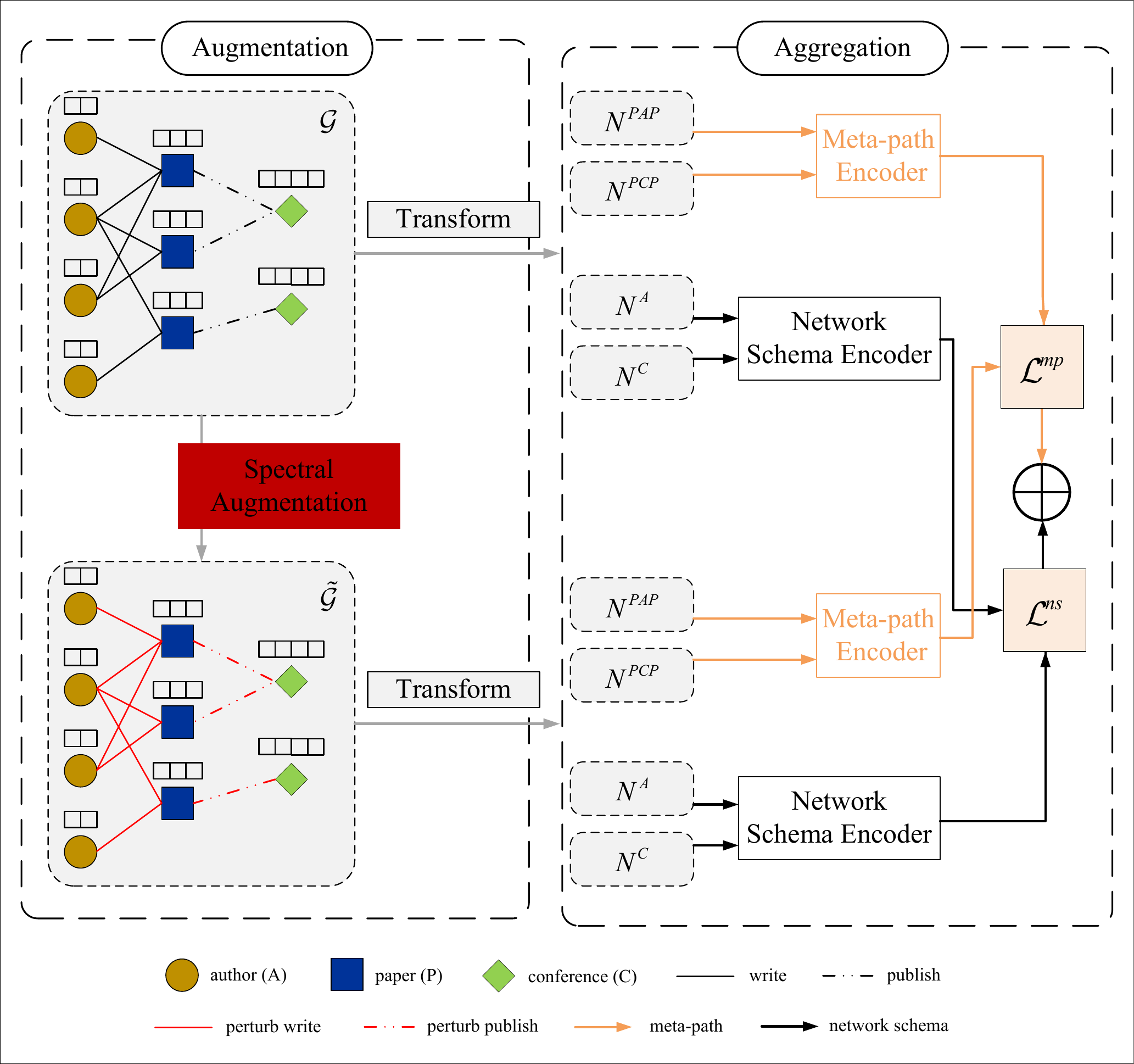}
	\caption{Illustration of the proposed SHCL.}
	\label{fig2}
\end{figure}
The proposed model SHCL uses spectral augmentation algorithm to generate multi-pair topology augmentation scheme, which can be used to explore the potential spectral invariance of heterogeneous graphs. The schematic diagram of the model framework is shown in Figure~\ref{fig2}. SHCL can be divided into two modules, including the augmentation module and the aggregation module. The first module includes spectral augmentation algorithm and topology augmentation scheme. Specifically, a pair of topological augmentation schemes are first parameterized, and then the spectral distance between the two schemes is maximized according to the spectral information of the heterogeneous graph itself, and the pair of topological augmentation schemes is optimized according to the spectral distance loss function. Then, according to the multi-pair topology augmentation scheme, the enhanced view with great difference in spectral distance between multiple pairs is obtained. The second module first generates a view representation of the two enhanced graphs using a double aggregation scheme, then calculates the weighted sum of the two losses based on the intra-scheme comparison, and finally learns the spectral invariance of the heterogeneous graphs.

\subsection{Parametric Topology Augmentation Scheme in Heterogeneous Graphs}
One of the difficulties of data augmentation on heterogeneous graphs is dealing with a variety of complex structures. For example, the adjacency matrix of a heterogeneous graph is asymmetric or extremely sparse, so it is difficult to calculate the Laplacian matrix of a heterogeneous graph directly. Therefore, in the process of spectral augmentation, additional processing is required for these special structures of heterogeneous graphs. We are inspired by the meta-path aggregation scheme in \citep{xie2023heterogeneous}. When using the meta-path aggregation scheme, the model generates multiple meta-path views based on the meta-path. These views are dense entitled undirected graphs, that is, the adjacency matrices of these views are real symmetric matrices, which means that it is easy to calculate the Laplacian matrix of these matrices, and further obtain the eigenvalues and eigenvectors of the Laplacian matrix. This subsection first defines a topological augmentation scheme determined by Bernoulli probability matrices. On this basis, SHCL formulates the objective of the augmentation scheme as maximizing the spectral distance between matrices.

The following describes the parameterized topology augmentation scheme. The topology augmentation scheme learned by SHCL mainly perturbs the meta-path view by modifying the weights in the meta-path view. For any meta-path view $G^\mathcal{P}$ determined by meta-path $\mathcal{P}$, we define a parameterized topology augmentation scheme $t(\textbf{A}^\mathcal{P})$ in this subsection, which controls the weight of the meta-path view through a Bernoulli probability matrix $\textbf{B}\in \{0,1\}^{n\times n}$, which can be formalized as follows:
\begin{equation}\label{eq1}
	t(\textbf{A}^\mathcal{P})=\textbf{A}^\mathcal{P}+\textbf{C}^\mathcal{P}\circ \textbf{B}^\mathcal{P}
\end{equation}	
where $\textbf{A}^\mathcal{P}$ is the adjacency matrix of the meta-path view $G^\mathcal{P}$, and $\textbf{C}^\mathcal{P}$ is the complement matrix of the adjacency matrix $\textbf{A}^\mathcal{P}$, which can be calculated by the following mathematical formula:
\begin{equation}\label{eq2}
	 C_{ij}^\mathcal{P} =
	\begin{cases}1 & \text{if } A_{ij}^\mathcal{P} = 0,\\
		A_{ij}^\mathcal{P} & \text{else. } 
	\end{cases} 
\end{equation}	

As can be seen from Eq(\ref{eq2}), for the weight edge that needs to be flipped, if its weight is $0$ in the adjacency matrix $\textbf{A}^\mathcal{P}$, it will be flipped to an edge with a weight of 1, otherwise it will be deleted. Note here that in general, the weights in the meta-path view are real numbers between 0 and 1, so when adding an edge between two nodes, the edge with weight 1 is created by default to disturb the meta-path view to the greatest extent. If the meta-path view of a heterogeneous graph does not meet the above prerequisites, it only needs to be normalized. Each element in $\textbf{B}^\mathcal{P}$ is drawn from the Bernoulli distribution with the corresponding weight in $\textbf{A}^\mathcal{P}$, i.e. $B_{ij}^\mathcal{P}\sim Bern(A_{ij}^\mathcal{P})$. In other words, if $B_{ij}^\mathcal{P}=1$, the corresponding weight edge needs to be flipped, and if $B_{ij}^\mathcal{P}=0$, the weight edge is directly retained.

Although a parameterized topology augmentation scheme has been defined, namely Eq(\ref{eq1}), the elements in the most critical Bernoulli probability matrix $\textbf{B}^\mathcal{P}$ have only two options of 0 and 1, which greatly reduces the learning ability of heterogeneous graph neural networks. And unlike the adjacency matrix of a homogeneous graph which is just a simple 0-1 matrix, the metapath view is a weighted undirected graph whose weight ranges from the whole set of real numbers, which is a natural advantage of heterogeneous graphs. Therefore, in this subsection, we take Bernoulli probability matrix $\textbf{B}^\mathcal{P}$ as the learning parameter of SHCL, and directly determine the values of each element in $\textbf{B}^\mathcal{P}$ by random gradient descent method, rather than sticking to the two choices of 0 and 1. In summary, the SHCL model can learn topological augmentation schemes specific to this heterogeneous graph by learning each parameter in $\textbf{B}^\mathcal{P}$.
\subsection{Spectral Augmentation Algorithms for Heterogeneous Graphs}
In the previous subsection, we defined a parameterized topology augmentation scheme, but how can a heterogeneous graph representation learning model design a topology augmentation scheme that best fits the heterogeneous graph itself? The spectral augmentation algorithm of heterogeneous graphs is introduced in detail below.

SHCL learns topological augmentation schemes of heterogeneous graphs from the perspective of spectrum. Specifically, the goal of the spectral augmentation algorithm is to find the matrix $\textbf{B}^\mathcal{P}$ that maximizes the spectral difference between the original and augmented graphs. Specifically, for the adjacency matrix $\textbf{A}^\mathcal{P}$ of the meta-path view $G^\mathcal{P}$, SHCL first computes the Laplacian matrix $\textbf{L}_\textbf{norm}^\mathcal{P}$ normalized by $\textbf{A}^\mathcal{P}$:
\begin{equation}
	\textbf{L}_\textbf{norm}^\mathcal{P}=Lap(\textbf{A}^\mathcal{P})=\textbf{I}_\textbf{n}-\textbf{D}^\textbf{-1/2}\textbf{A}^\mathcal{P}\textbf{D}^\textbf{-1/2}
\end{equation}	
where $\textbf{I}_\textbf{n}$ is the $n\times n$ identity matrix. In order to find the graph spectral of the adjacency matrix $\textbf{A}^\mathcal{P}$, it is necessary to decompose the features of the Laplacian matrix $\textbf{L}_\textbf{norm}^\mathcal{P}$, which can be formalized as the following formula:
\begin{equation}
	\textbf{L}_\textbf{norm}^\mathcal{P}=\textbf{U}\mathbf{\Lambda}^\mathcal{P}\textbf{U}^\top
\end{equation}	
The diagonal matrix $\mathbf{\Lambda}^\mathcal{P}=eig(\textbf{L}_\textbf{norm}^\mathcal{P})=eig(Lap(\textbf{A}^\mathcal{P}))=diag(\lambda_1,...,\lambda_n)$ consists of real eigenvalues, which are called the Graph Spectrum. Based on the above concepts, this subsection can define the spectral distance $\Delta_{\text{single}}$ of the augmented graph and the original graph:
\begin{equation}
\Delta_{\text{single}}=\Big\|eig(Lap(\textbf{A}^\mathcal{P}+\textbf{C}^\mathcal{P}\circ \textbf{B}^\mathcal{P}))-eig(Lap(\textbf{A}^\mathcal{P}))\Big\|
\end{equation}	

According to the concept of spectral distance, we can further define the learning objectives of the spectral augmentation algorithm of the original graph and the single augmented graph:
\begin{equation}
	L_{\text{single}}=\Big\|eig(Lap(\textbf{A}^\mathcal{P}+\textbf{C}^\mathcal{P}\circ \textbf{B}^\mathcal{P}))-eig(Lap(\textbf{A}^\mathcal{P}))\Big\|
\end{equation}	
We can also derive the following two variants based on the goal of maximizing spectral distance:
\begin{equation}\label{eqmax}
	L_{\text{single}}^{\max}=\max \frac{\Big\|eig(Lap(\textbf{A}^\mathcal{P}+\textbf{C}^\mathcal{P}\circ \textbf{B}^\mathcal{P}))\Big\|}{\Big\|eig(Lap(\textbf{A}^\mathcal{P}))\Big\|}
\end{equation}	
\begin{equation}\label{eqmin}
	L_{\text{single}}^{\min}=\min \frac{\Big\|eig(Lap(\textbf{A}^\mathcal{P}+\textbf{C}^\mathcal{P}\circ \textbf{B}^\mathcal{P}))\Big\|}{\Big\|eig(Lap(\textbf{A}^\mathcal{P}))\Big\|}
\end{equation}	

The above three objective functions can improve the spectral difference between the augmented graph and the original graph, but Eq(\ref{eqmax}) and Eq(\ref{eqmin}) can control the amplitude of the augmented graph change according to the graph information of the original graph. This can be further extended to maximize the spectral distance $\Delta_{\text{double}}$ between the two augmented graphs:
\begin{equation}
	\Delta_{\text{double}}=\Big\|eig(Lap(\textbf{A}^\mathcal{P}+\textbf{C}^\mathcal{P}\circ \textbf{B}_1^\mathcal{P}))-eig(Lap(\textbf{A}^\mathcal{P}+\textbf{C}^\mathcal{P}\circ \textbf{B}_2^\mathcal{P}))\Big\|
\end{equation}	

To maximize the spectral distance $\Delta_{\text{double}}$, the following three objective functions can be defined:
\begin{equation}
	L_{\text{double}}=\max\Big\|eig(Lap(\textbf{A}^\mathcal{P}+\textbf{C}^\mathcal{P}\circ \textbf{B}_1^\mathcal{P}))-eig(Lap(\textbf{A}^\mathcal{P}+\textbf{C}^\mathcal{P}\circ \textbf{B}_2^\mathcal{P}))\Big\|
\end{equation}	
\begin{equation}\label{eqmaxl}
		L_{\text{double}}^{\max}=\max\frac{\Big\|eig(Lap(\textbf{A}^\mathcal{P}+\textbf{C}^\mathcal{P}\circ \textbf{B}_1^\mathcal{P}))\Big\|}{\Big\|eig(Lap(\textbf{A}^\mathcal{P}+\textbf{C}^\mathcal{P}\circ \textbf{B}_2^\mathcal{P}))\Big\|}
\end{equation}	
\begin{equation}\label{eqminl}
	L_{\text{double}}^{\min}=\min\frac{\Big\|eig(Lap(\textbf{A}^\mathcal{P}+\textbf{C}^\mathcal{P}\circ \textbf{B}_1^\mathcal{P}))\Big\|}{\Big\|eig(Lap(\textbf{A}^\mathcal{P}+\textbf{C}^\mathcal{P}\circ \textbf{B}_2^\mathcal{P}))\Big\|}
\end{equation}	
It should be noted here that since SHCL only needs to maximize the spectral distance between the augmented graphs. In essence, Eq(\ref{eqmaxl}) and Eq(\ref{eqminl}) are the same. So it is useful to substitute Eq(\ref{eqmaxl}) for Eq(\ref{eqminl}).

Because there are two learnable parameters in the objective function, the above two objective functions are difficult to optimize directly. The Eq(\ref{eqmaxl}) can be modified as follows:
\begin{equation}
	\begin{aligned}
	L_{\text{double}}^{\max}&=\max\frac{\Big\|eig(Lap(\textbf{A}^\mathcal{P}+\textbf{C}^\mathcal{P}\circ \textbf{B}_1^\mathcal{P}))\Big\|\times\Big\|eig(Lap(\textbf{A}^\mathcal{P})\Big\|}{\Big\|eig(Lap(\textbf{A}^\mathcal{P})\Big\|\times\Big\|eig(Lap(\textbf{A}^\mathcal{P}+\textbf{C}^\mathcal{P}\circ \textbf{B}_2^\mathcal{P}))\Big\|}\\&=\max \left(\frac{\Big\|eig(Lap(\textbf{A}^\mathcal{P}+\textbf{C}^\mathcal{P}\circ \textbf{B}_1^\mathcal{P}))\Big\|}{\Big\|eig(Lap(\textbf{A}^\mathcal{P})\Big\|}\times\frac{\Big\|eig(Lap(\textbf{A}^\mathcal{P})\Big\|}{\Big\|eig(Lap(\textbf{A}^\mathcal{P}+\textbf{C}^\mathcal{P}\circ \textbf{B}_2^\mathcal{P}))\Big\|} \right)\\&=\max \frac{\Big\|eig(Lap(\textbf{A}^\mathcal{P}+\textbf{C}^\mathcal{P}\circ \textbf{B}_1^\mathcal{P}))\Big\|}{\Big\|eig(Lap(\textbf{A}^\mathcal{P})\Big\|}\times\min\frac{\Big\|eig(Lap(\textbf{A}^\mathcal{P}+\textbf{C}^\mathcal{P}\circ \textbf{B}_2^\mathcal{P}))\Big\|}{\Big\|eig(Lap(\textbf{A}^\mathcal{P})\Big\|}\\&=\frac{\left(\max\Big\|eig(Lap(\textbf{A}^\mathcal{P}+\textbf{C}^\mathcal{P}\circ \textbf{B}_1^\mathcal{P}))\Big\|\right)\left(\min\Big\|eig(Lap(\textbf{A}^\mathcal{P}+\textbf{C}^\mathcal{P}\circ \textbf{B}_2^\mathcal{P}))\Big\|\right)}{\Big\|eig(Lap(\textbf{A}^\mathcal{P})\Big\|_2^2}
    \end{aligned}
\end{equation}	

Therefore, the complex optimization problem of Eq(\ref{eqmaxl}) can be decomposed into two simple optimization problems: 1) Maximizing the graph of the first augmented graph; 2) Minimize the graph of the second augmented graph. Namely:
\begin{equation}
	\max\frac{\Big\|eig(Lap(\textbf{A}^\mathcal{P}+\textbf{C}^\mathcal{P}\circ \textbf{B}_1^\mathcal{P}))\Big\|}{\Big\|eig(Lap(\textbf{A}^\mathcal{P})\Big\|}
\end{equation}	
\begin{equation}
	\max\frac{\Big\|eig(Lap(\textbf{A}^\mathcal{P})\Big\|}{\Big\|eig(Lap(\textbf{A}^\mathcal{P}+\textbf{C}^\mathcal{P}\circ \textbf{B}_2^\mathcal{P}))\Big\|}
\end{equation}	
It should be noted here that the optimization goal of SHCL still retains a form similar to Eq(\ref{eqmax}), that is, the graph information of the original graph is retained in the denominator (or numerator), in order to better control the degree of graph changes of the enhanced graph. In this section, we argue that if the patterns of two augmented graphs are too different, the contrastive learning model may not learn any spectral invariance. Therefore, the objective function of the SHCL model in learning the first topology augmentation scheme is:
\begin{equation}
	\mathcal{J}_{up}=\max\frac{\Big\|eig(Lap(\textbf{A}^\mathcal{P}+\textbf{C}^\mathcal{P}\circ \textbf{B}_1^\mathcal{P}))\Big\|}{\Big\|eig(Lap(\textbf{A}^\mathcal{P})\Big\|}
\end{equation}	
The objective function of learning the second topology augmentation scheme is as follows:
\begin{equation}
	\mathcal{J}_{down}=\max\frac{\Big\|eig(Lap(\textbf{A}^\mathcal{P})\Big\|}{\Big\|eig(Lap(\textbf{A}^\mathcal{P}+\textbf{C}^\mathcal{P}\circ \textbf{B}_2^\mathcal{P}))\Big\|}
\end{equation}	

Finally, when SHCL learns the first or second topology augmentation scheme, $\textbf{B}^\mathcal{P}$ initializes all elements to 0 by default. This can prevent false optimization problems after random initialization. For example, when learning the first topology augmentation scheme, the $\textbf{B}_1^\mathcal{P}$ of random initialization may cause the initial value of the formula to be less than 1, so the model needs to consume more resources in the topology augmentation scheme. The same principle applies to $\textbf{B}_2^\mathcal{P}$.

In summary, SHCL learns two augmentation scheme control matrices $\textbf{B}_1^\mathcal{P}$ and $\textbf{B}_2^\mathcal{P}$ according to meta-path view $G^\mathcal{P}$, and then can obtain the adjacency matrices $\mathbf{A}_1^\mathcal{P}$ and $\textbf{A}_2^\mathcal{P}$ of two meta-path augmentation views $G_1^\mathcal{P}$ and $G_2^\mathcal{P}$ respectively. For $M$ meta-paths, a heterogeneous graph has $M$ meta-path views $\left\{G^{\mathcal{P}_1},...,G^{\mathcal{P}_M}\right\}$. So SHCL can calculate two meta path augmentation view sets $\gamma_1=\left\{G_1^{\mathcal{P}_1},...,G_1^{\mathcal{P}_M}\right\}$ and $\gamma_2=\left\{G_2^{\mathcal{P}_1},...,G_2^{\mathcal{P}_M}\right\}$. It should be noted that the optimization and learning of these augmentation schemes only need to be done once. And the heterogeneous graph contrastive learning model does not need to be calculated again when optimizing parameters. So it does not bring any additional complexity to the contrastive learning process. And the spectral augmentation algorithm proposed in this section can be combined with any other heterogeneous graph contrastive learning model, such as the contrastive learning model on heterogeneous graphs (CLHG) proposed in \citep{xie2023heterogeneous}.

Finally, SHCL uses the meta-path aggregation scheme proposed in \citep{xie2023heterogeneous} to aggregate the higher-order semantic information of the augmented view sets of the two meta-paths respectively, and assists the SHCL model to learn the spectral invariance of heterogeneous graphs by comparing within the scheme.

\section{Experiments}\label{sec:exp}
In this section, we verify the performance of the SHCL model. We tested the results of the node classification task on three real-world data sets and compared them with eleven classical and advanced related methods. In the ablation experiment, the rationality of spectral augmentation algorithm design and the effectiveness of modeling spectral invariance of heterogeneous information were verified. The following is an introduction.

\subsection{Datasets}
To validate the model proposed in this paper, the experiment uses the following three real-world datasets: the academic datasets DBLP\footnote[1]{ https://dblp.org/.} and ACM\footnote[2]{https://dl.acm.org/.}, and the film dataset Freebase\footnote[3]{https://www.freebase.com/.}. The basic statistics of the datasets are shown in Table \ref{tabdataset}, and the details are described as follows:

\begin{table}
	\caption{Statistics of the datasets.}
	\centering
	\renewcommand\arraystretch{1.5}
	\resizebox{\linewidth}{!}{
	\begin{tabular}{cccccc}
		\toprule
		\textbf{Dataset} & \textbf{Type of node} & \textbf{Number of nodes} & \textbf{Type of edge} & \textbf{Number of edges} & \textbf{Meta-path} \\ 
		\hline
		\multirow{3}*{DBLP} & \multirow{3}*{\makecell[c]{Author (A) \\ Paper (P) \\ Conference (C) \\Term (T)}} & \multirow{3}*{\makecell[c]{4057 \\ 14328 \\ 20 \\ 7723}} & P-A & 19645 & APA \\
		\cline{4-6}
			& & & P-C & 14328 & APCPA \\
		\cline{4-6} 
			& & & P-T & 85810 & APTPA \\
		\hline
		\multirow{2}*{ACM}& \multirow{2}*{\makecell[c]{Author (A) \\ Paper (P) \\ Subject (S) \\}} & \multirow{2}*{\makecell[c]{4019 \\ 7167 \\ 60 \\}} & P-A & 13407 & PAP \\
		\cline{4-6}
		& & & P-S & 4019 & PSP \\
		\hline
		\multirow{3}*{Freebase} & \multirow{3}*{\makecell[c]{Movie (M) \\ Actor (A) \\ Direct (D) \\Writer (W)}} & \multirow{3}*{\makecell[c]{3492 \\ 33401 \\ 2502 \\ 4495}} & M-A & 65341 & MAM\\
		\cline{4-6}
		& & & M-D & 3762 & MDM \\
		\cline{4-6} 
		& & & M-W & 6414 & MWM \\
		\bottomrule
	\end{tabular}}
	\label{tabdataset}
\end{table}
\begin{itemize}
	\item \textit{DBLP Dataset} \citep{fu2020magnn} comes from the DBLP Academic Search website, where the node set includes 4057 authors, 14328 papers, 20 conferences, and 7723 terms. The relationship (edge) set consists of 19645 author-paper (P-A) edges, 14328 author-conference (P-C) edges, and 85810 author-term (P-T) edges. The classification target of the node classification task in this dataset is author, which is divided into four categories. The meta-path-dependent approach uses three meta-paths, namely APA, APCPA, and APTPA.
	\item \textit{ACM Dataset} \citep{zhao2020network} extracts the network of papers from the ACM database, where the node set includes 7167 papers, 4019 authors, and 60 subjects. The relationship (edge) set consists of 13,407 paper - author (P-A) edges and 4,019 paper - subject (P-S) edges. The classification target of the node classification task in this dataset is paper, which is divided into three categories. The meta-path-dependent approach uses two meta-paths, PAP and PSP.
	\item \textit{Freebase dataset} \citep{li2021leveraging} is a Movie dataset from the Freebase website, where the node set includes 3492 movies, 33,401 actors, 2502 directors and 4459 writers. The relationship (edge) set consists of 65,341 movie - actor (M-A) edges, 3,762 movie - director (M-D) edges, and 6,414 movie - writer (M-W) edges. The classification target of the node classification task in this dataset is movie, which are divided into three categories. The metapath dependent approach uses three meta-paths, MAM, MDM, and MWM.
\end{itemize}

\subsection{Experimental Setup}
\textbf{Baseline.} In order to comprehensively verify the performance of SHCL, we consider comparison with four major categories of methods, including:  \romannumeral1) Unsupervised homogenous graph represent learning methods GraphSAGE \citep{hamilton2017inductive}, VGAE \citep{kipf2016variational} and DGI \citep{velickovic2019deep};  \romannumeral2) Unsupervised heterogeneous graph represent learning methods  Metapath2Vec \citep{dong2017metapath2vec}, HERec \citep{shi2018heterogeneous}, HetGNN \citep{ren2019heterogeneous} and DMGI \citep{park2020unsupervised};  \romannumeral3) Semi-supervised heterogeneous graph represent learning method HAN \citep{wang2019heterogeneous}. \romannumeral4) Heterogeneous graph contrastive learning methods HeCo \citep{wang2021self} and CLHG \citep{xie2023heterogeneous}. The differences in these comparison methods are summarized below:
\begin{itemize}
	\item \textbf{GraphSAGE} \citep{hamilton2017inductive} is a homogeneous graph represent learning framework based on convolutional graph neural networks, which can be applied to induction scenarios of dynamic graphs.
	\item \textbf{VGAE} \citep{kipf2016variational} is a variant model that applies variational autoencoders to a homogeneous graph structure.
	\item \textbf{DGI} \citep{velickovic2019deep} is a classical technique for unsupervised graph contrastive learning, which utilizes Infomax theory for homogeneous graph contrastions.
	\item \textbf{Metapath2Vec} \citep{dong2017metapath2vec} is a heterogeneous graph represent learning method that uses meta-paths, but it can only take one meta-path as input.
	\item \textbf{HERec} \citep{shi2018heterogeneous} is a heterogeneous network recommendation method based on heterogeneous graph node represent learning.
	\item \textbf{HetGNN} \citep{ren2019heterogeneous} is an improved GNNs method that can take advantage of heterogeneous information about node and edge types.
	\item \textbf{DMGI} \citep{park2020unsupervised} is an unsupervised node represent learning method for learning heterogeneous networks.
	\item \textbf{HAN} \citep{wang2019heterogeneous} is a semi-supervised heterogeneous graph represent learning method that uses hierarchical attention to automatically distinguish the importance of meta-paths.
	\item \textbf{HeCo} \citep{wang2021self} is a heterogeneous graph contrastive learning method under unsupervised conditions, which employs a cross-view contrastive scheme to learn node representations.
	\item \textbf{CLHG} \citep{xie2023heterogeneous} is a contrastive learning method for heterogeneous graphs. It adopts adaptive augmentation algorithm and double aggregation scheme in spatial topology to model higher-order semantic information and local heterogeneous structure information of heterogeneous graphs, but it does not examine the key spectral invariance in contrastive learning of heterogeneous graphs from the perspective of spectrum.
\end{itemize}

\textbf{Evaluation Metrics.} We use three widely-used metrics \textit{Macro F1-score} (MaF1), \textit{Micro F1-score} (Mi-F1), and \textit{Area Under Curve} (AUC) to evaluate the model. F1 score is an indicator used in statistics to measure the accuracy of a binary model, which also considers the accuracy and recalls of the classification model \citep{tong2018emotion}.

For the experimental setup of the comparison method, the experiment followed the setup described in the original paper as well as the setup in HeCo \citep{wang2021self}. 

For the SHCL model proposed in this paper, in all datasets, the optimizer is Adam algorithm \citep{zhang2019heterogeneous}, the contrastive learning rate is 0.001, and the embedding dimension is set to 64. The spectral augmentation algorithm learned different datasets at inconsistent learning rates. And the experiment tested values from 0.01 to 0.2 in increments of 0.01. Finally, the optimal learning rate of DBLP dataset is 0.1, that of ACM dataset is 0.07, and that of Freebase dataset is 0.09. In addition, the experiment performed a series of tests to determine the optimal values of the hyperparameters $P_\tau$, $P_e$, $\tau$ and $\lambda$. For the $P_\tau$ of the adaptive augmentation scheme, the experiment tests values from 0.0 to 1.0 in increments of 0.1. For $P_e$, the experiment tested values from 0.0 to 1.0 in increments of 0.1. For $\tau$, the experiment tests values from 0.1 to 0.9 in increments of 0.1. Similarly, for $\lambda$ in the final loss function, the experiment tests values from 0.0 to 1.0 in increments of 0.1. Based on the test results, the optimal values of the hyperparameters $P_\tau$, $P_e$, $\tau$ and $\lambda$ are 0.7, 0.7, 0.5 and 0.5, respectively. In the shard of the datasets, all the shard test sets and validation sets each contain 1000 nodes. The only difference is that 20, 40 and 60 marked nodes are selected for each class as the training sets of the different shard datasets. Run 10 times at random and report the average result and standard deviation.

\subsection{Experimental Results and Discussion}
This subsection presents the experimental results on different types of datasets and discuss them.

\subsubsection{Results of Comparison Experiment}
\begin{table}
	\caption{Performance comparison of all methods on DBLP dataset for node classification task}
	\centering
	\renewcommand\arraystretch{1.5}
	\begin{threeparttable}
	\resizebox{\linewidth}{!}{
	\begin{tabular}{cccccccccc} 
		\toprule
		\multirow{2}*{Methods} &  \multicolumn{3}{c}{\textbf{Macro-F1}$\mathbf{(\%)}$}  & \multicolumn{3}{c}{\textbf{Micro-F1}$\mathbf{(\%)}$}  & \multicolumn{3}{c}{\textbf{AUC}$\mathbf{(\%)}$} \\
		\cline{2-10}
		& 20 & 40 & 60 & 20 & 40 & 60 & 20 & 40 & 60 \\
		\hline
		GraphSAGE & 71.97 & 73.69 & 73.86 & 71.44 & 73.61 & 74.05 & 90.59 & 91.42 & 91.73 \\
		VGAE & 90.90 & 89.60 & 90.08 & 91.55 & 90.00 &  90.95 & 98.15 & 97.85 & 98.37 \\
		Metapath2vec & 88.98 & 88.68 & 90.25 & 89.67 & 89.14 & 91.17 & 97.69 & 97.08 & 98.00 \\
		HERec & 89.57 & 89.73 & 90.18 & 90.24 & 90.15 &  91.01 & 98.21 & 97.93 & 98.49 \\
		HetGNN & 89.51 & 88.61 & 89.56 & 90.11 & 89.03 & 90.43 & 97.96 & 97.70 & 97.97 \\
		HAN & 89.31 & 88.87 & 89.20 & 90.16 & 89.47 & 90.34 & 98.07 & 97.48 & 97.96 \\
		DGI & 87.93 & 88.62 & 89.19 & 88.72 & 89.22 & 90.35 & 96.99 & 97.12 & 97.76 \\
		DMGI & 89.94 & 89.25 & 89.46 & 90.78 & 89.92 & 90.66 & 97.75 & 97.23 & 97.72 \\
		HeCo & 91.04 & 90.11 & 90.61 & 91.76 & 90.54 & 91.59 & 98.28 & 98.4 & 98.65 \\
		CLHG & \textbf{92.58} & 91.08 & 91.39 & \textbf{93.10} & 91.42 & 92.12 & \textbf{98.47} & 98.16 & \textbf{98.69} \\
		\textbf{SHCL} & 92.45 & \textbf{91.49} & \textbf{91.76} & 93.02 & \textbf{91.81} & \textbf{92.50} & 98.43 & \textbf{98.19} & 98.52 \\
		\bottomrule
	\end{tabular}}
	\label{tabdblp}
	 \begin{tablenotes}
		\footnotesize
		\item Note: Optimal results are bolded.
	 \end{tablenotes}
	\end{threeparttable}
\end{table}

\begin{table}
	\caption{Performance comparison of all methods on ACM dataset for node classification task}
	\centering
	\renewcommand\arraystretch{1.5}
	\begin{threeparttable}
		\resizebox{\linewidth}{!}{
			\begin{tabular}{cccccccccc} 
				\toprule
				\multirow{2}*{Methods} &  \multicolumn{3}{c}{\textbf{Macro-F1}$\mathbf{(\%)}$}  & \multicolumn{3}{c}{\textbf{Micro-F1}$\mathbf{(\%)}$}  & \multicolumn{3}{c}{\textbf{AUC}$\mathbf{(\%)}$} \\
				\cline{2-10}
				& 20 & 40 & 60 & 20 & 40 & 60 & 20 & 40 & 60 \\
				\hline
				GraphSAGE & 47.13 & 55.96 & 56.59 & 49.72 & 60.98 & 60.72 & 65.88 & 71.06 & 70.45 \\
				VGAE & 62.72 & 61.61 & 61.67 & 68.02 & 66.38 & 65.71 & 79.50 & 79.14 & 77.90 \\
				Meapath2vec & 51.91 & 62.41 & 61.13 & 53.13 & 64.43 & 62.72 & 71.66 & 80.48 & 79.33 \\
				HERec & 55.13 & 61.21 & 64.35 & 57.47 & 62.62 & 65.15 & 75.44 & 79.84 & 81.64 \\
				HetGNN & 72.11 & 72.02 & 74.33 & 71.89 & 74.46 & 76.08 & 84.36 & 85.01 & 87.64 \\
				HAN & 85.66 & 87.47 & 88.41 & 85.11 & 87.21 & 88.10 & 93.47 & 94.84 & 94.68 \\
				DGI & 79.27 & 80.23 & 80.03 & 76.63 & 80.41 & 80.15 & 91.47 & 91.52 & 91.41 \\
				DMGI & 87.86 & 86.23 & 87.97 & 87.60 & 86.02 & 87.82 & 96.72 & 96.35 & 96.79 \\
				HeCo & 88.05 & 87.38 & 89.09 & 87.74 & 87.09 & 88.82 & 96.29 & 96.12 & 96.41 \\
				CLHG & 88.90 & 90.16 & 89.11 & 88.64 & 89.91 & 88.97 & 97.51 & \textbf{97.99} & \textbf{97.40} \\
				\textbf{SHCL} & \textbf{89.61} & \textbf{90.28} & \textbf{89.32} & \textbf{89.67} & \textbf{90.33} & \textbf{89.32} & \textbf{97.53} & 97.86 & 96.72 \\
				\bottomrule
		\end{tabular}}
		\label{tabacm}
		\begin{tablenotes}
			\footnotesize
			\item Note: Optimal results are bolded.
		\end{tablenotes}
	\end{threeparttable}
\end{table}

\begin{table}
	\caption{Performance comparison of all methods on Freebase dataset for node classification task}
	\centering
	\renewcommand\arraystretch{1.5}
	\begin{threeparttable}
		\resizebox{\linewidth}{!}{
			\begin{tabular}{cccccccccc} 
				\toprule
				\multirow{2}*{Methods} &  \multicolumn{3}{c}{\textbf{Macro-F1}$\mathbf{(\%)}$}  & \multicolumn{3}{c}{\textbf{Micro-F1}$\mathbf{(\%)}$}  & \multicolumn{3}{c}{\textbf{AUC}$\mathbf{(\%)}$} \\
				\cline{2-10}
				& 20 & 40 & 60 & 20 & 40 & 60 & 20 & 40 & 60 \\
				\hline
				GraphSAGE & 45.14 & 44.88 & 45.16 & 54.83 & 57.08 & 55.92 & 67.63 & 66.42 & 66.78 \\
				VGAE & 53.81 & 52.44 & 50.65 & 55.20 & 56.05 & 53.85 & 73.03 & 74.05 & 71.75 \\
				Meapath2vec & 53.96 & 57.80 & 55.94 & 56.23 & 60.01 & 58.74 & 71.78 & 75.51 & 74.78 \\
				HERec & 55.78 & 59.28 & 56.50 & 57.92 & 62.71 & 58.57 & 73.89 & 76.08 & 74.89 \\
				HetGNN & 52.72 & 48.57 & 52.37 & 56.85 & 53.96 & 56.84 & 70.84 & 69.48 & 71.01 \\
				HAN & 53.16 & 59.63 & 56.77 & 57.24 & 63.74 & 61.06 & 73.26 & 77.74 & 75.68 \\
				DGI & 54.90 & 53.40 & 53.81 & 58.16 & 57.82 & 67.96 & 72.80 & 72.97 & 73.32 \\
				DMGI & 55.79 & 49.88 & 53.10 & 58.26 & 54.28 & 56.69 & 73.19 & 70.77 & 73.17 \\
				HeCo & 58.38 & \textbf{61.20} & 60.48 & 60.91 & 64.11 & 64.11 & 75.36 & \textbf{78.71} & 78.10 \\
				CLHG & \textbf{69.91} & 60.63 & 60.91 & \textbf{60.12} & 63.14 & 63.83 & \textbf{78.27} & 78.14 & 78.32 \\
				\textbf{SHCL} & 58.49 & 60.87 & \textbf{61.03} & 61.37 & \textbf{64.33} & \textbf{64.34} & 76.88 & 77.85 & \textbf{78.68} \\
				\bottomrule
		\end{tabular}}
		\label{tabfb}
		\begin{tablenotes}
			\footnotesize
			\item Note: Optimal results are bolded.
		\end{tablenotes}
	\end{threeparttable}
\end{table}
The node classification results of SHCL model on DBLP, ACM and Freebase data sets are shown in Table \ref{tabdblp}, Table \ref{tabacm} and Table \ref{tabfb} respectively. By observing the node classification results of different methods, we have the following analysis conclusions:
\begin{enumerate}[(i)]
	\item Compared to HeCo, the most advanced method, the SHCL model achieves the best performance on almost all shards of the three datasets, as well as on all metrics. In the DBLP dataset, the average performance of SHCL on Macro-F1, Micro-F1 and AUC is $1.45\%$, $1.26\%$ and $0.06\%$ higher than that of HeCo, respectively. In the ACM dataset, the average performance of SHCL on Macro-F1, Micro-F1 and AUC is $1.77\%$, $2.15\%$ and $1.14\%$ higher than HeCo, respectively. Meanwhile, in the Freebase dataset, the average performance of SHCL on Macro-F1, Micro-F1 and AUC improved by $0.18\%$, $0.48\%$ and $0.53\%$, respectively, compared to HeCo. And on both the DBLP and ACM datasets, SHCL achieves the best performance on average across all methods.The above results strongly demonstrate the promise of the proposed heterogeneous graph augmentation algorithm.
	\item SHCL has more of a boost on the ACM dataset than on the DBLP dataset. The possible reason for this is that all edges in the ACM data set are associated with the target node. The higher-order semantic information of the target node can better reflect the potential semantic of the complete heterogeneous graph, which is more conducive to the topology augmentation scheme designed by spectral augmentation algorithm to change the graph information of the meta-path view. Finally, SHCL is able to model spectral invariance in heterogeneous graphs.
	\item In comparison with the newly proposed model, CLHG, SHCL lags slightly in a small number of test results, but the average performance of SHCL still exceeds that of CLHG in Macro-F1 and Micro-F1. Specifically, it is improved by $0.24\%$ and $0.25\%$ on DBLP dataset respectively. On the ACM dataset, the increase is $0.39\%$ and $0.67\%$, respectively. These experimental results can strongly prove the prospect of using spectral augmentation algorithm to model the spectral invariance of heterogeneous graphs.
	\item It is also observed that SHCL is inferior to CLHG in some shards of the Freebase dataset. This is due to the lack of characteristics of target type nodes in the Freebase dataset. This shows that the spectral invariance of the heterogeneous graph is related to the characteristics of the target node, which indicates that the spectral invariance of the heterogeneous graph contains some node characteristic information. This further indicates that spectral invariance, which should be modeled by the contrastive learning model proposed in this paper, is the essential information in heterogeneous graphs, and the rich information of heterogeneous graphs modeled from the spectrum dimension is more important.
\end{enumerate}

\subsubsection{Ablation experiment}
To demonstrate the effectiveness of the spectral augmentation algorithm design in the SHCL model, we conducted several sets of ablation experiments under the node classification task. And we also further analyzed the contribution of the spectral augmentation algorithm in the SHCL. The following two model variants were considered for ablation experiments:
\begin{itemize}
	\item \textbf{SHCL-S}: The meta-path augmented view is generated by directly copying the original meta-path augmented view without using the topology augmentation scheme learned by any spectral augmentation algorithm.
	\item \textbf{SHCL-L}: The control matrix $\textbf{B}^\mathcal{P}$  of the topology augmentation scheme is not learned by spectral augmentation algorithm, but the matrix elements are randomly initialized. Then use the randomly initialized topology augmentation scheme directly to generate the meta-path augmented view.
\end{itemize}
\begin{figure}
	\centering
	\includegraphics[width=5.4in]{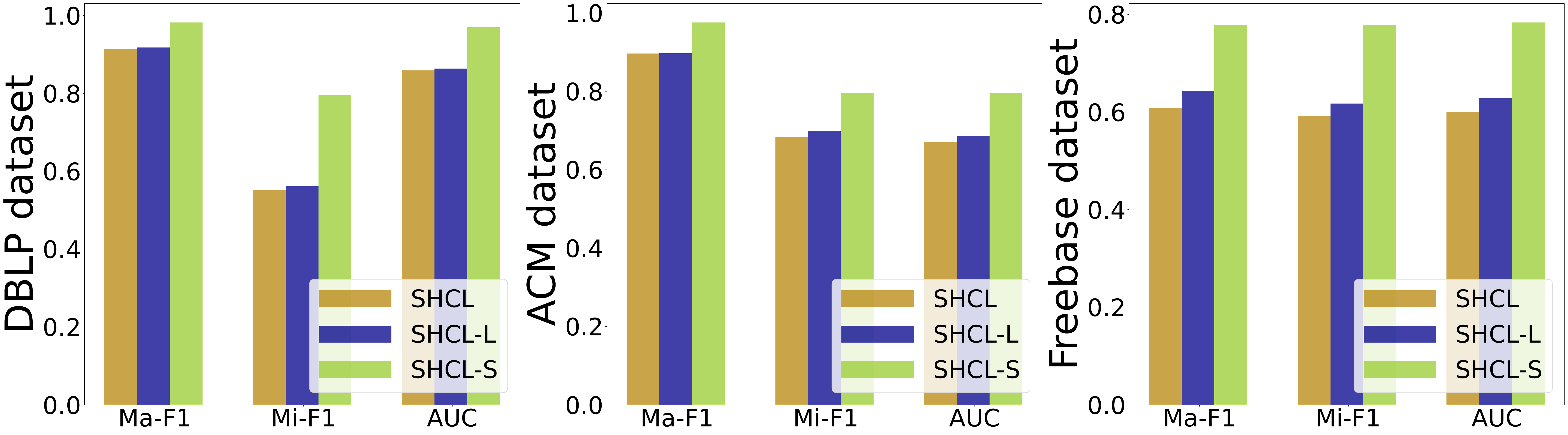}
	\caption{Comparison of different SHCL model variants.}
	\label{fig3}
\end{figure}
We compare these model variants with the SHCL model on all datasets and report results on three evaluation metrics, Macro-F1, Micro-F1, and AUC, respectively. The results of the ablation experiment are shown in Figure \ref{fig3}. We have the following observations:
\begin{enumerate}[(i)]
	\item The model variant SHCL-S outperforms SHCL-L on both the DBLP and Freebase datasets. This non-directional randomly generated topology augmentation scheme is likely to degrade the performance of heterogeneous graph contrastive learning models. In other words, the spectral invariance of heterogeneous graphs cannot be mined effectively, which further indicates that it is feasible to model the invariance of heterogeneous graphs by maximizing the spectral distance between views.
	\item Model variant SHCL-S only lags behind SHCL-L in the ACM dataset, which may be due to the fact that the randomly generated topology augmentation scheme just expands the spectral information difference of the augmented view, thus improving the ability of model variant to mine the invariance of heterogeneous graphs.
	\item Both SHCL-S and SHCL-L are inferior to the SHCL model on all datasets and all indicators. This shows that the spectral augmentation algorithm of heterogeneous graphs successfully enables SHCL to mine the spectral invariance of heterogeneous graphs by maximizing the spectral distance between augmented views, which further indicates that the spectral augmentation algorithm has broad prospects in the use of heterogeneous graphs.
\end{enumerate}

\section{Conclusion}\label{sec:con}
In this paper, we first propose a spectral augmentation algorithm for heterogeneous graphs, and further propose a contrastive learning model for heterogeneous graphs, called SHCL. The main objective of the model is to learn multiple pairs of meta-path view specific topological augmentation schemes based on the spectral information of the meta-path view, which can disturb the spectral information of heterogeneous graphs to the greatest extent from the spectral perspective. SHCL then learns the potential spectral invariance of the heterograph from these augmented views with large spectral differences. The biggest advantage of spectral augmentation algorithm is that it does not need any prior knowledge, and can learn the best topology augmentation scheme according to the information carried by the heterogeneous graph itself. Then, SHCL uses a dual aggregation scheme to aggregate these important semantic information and mine the potential spectral invariance among them. A large number of experiments were conducted on three publicly available and real datasets. The experimental results show that SHCL is significantly superior to the most advanced methods.

For the problem of how to model the most essential information in heterogeneous graphs, although the spectral augmentation algorithm is proposed in this paper from the spectrum perspective to assist the model to model the spectral invariance of heterogeneous graphs, the most essential information of heterogeneous graphs is still not mined from the most intuitive spatial topological level. How to explore and mine the essential information of heterogeneous graphs in spatial topology is also an important and key research direction.


%
%


\bibliography{lnc}

\end{document}